\documentclass{article}

\usepackage{arxiv}

\usepackage[utf8]{inputenc} 
\usepackage[T1]{fontenc}    
\usepackage{hyperref}       
\usepackage{url}            
\usepackage{booktabs}       
\usepackage{amsfonts}       
\usepackage{nicefrac}       
\usepackage{microtype}      
\usepackage{lipsum}

 \usepackage{graphicx}
\usepackage{float}
\usepackage{rotating}
\usepackage[]{xcolor}
\usepackage{subcaption}

\title{\color{black}MNIST-NET10: A heterogeneous deep  networks fusion based on the degree of certainty to reach 0.1 error rate. Ensembles overview and proposal}

\author{
  Siham Tabik\\
  Andalusian Research Institute\\ in Data Science and\\ Computational Intelligence\\ University of Granada 18071 
  Spain\\ 
  \texttt{siham@ugr.es} \\
\And
Ricardo F. Alvear-Sandoval\\
GAMMA-L+/DTSC\\ Universidad Carlos III de Madrid\\
Spain\\ 
  \texttt{r.f.alvear.s@gmail.com} \\
\And
María M. Ruiz\\
  Andalusian Research Institute\\ in Data Science and\\ Computational Intelligence\\  University of Granada\\ 18071 
  Spain\\ 
  \texttt{mariadelmaruiz25@gmail.com} \\
\And
José-Luis Sancho-Gómez\\
TDAM/DTIC\\ Universidad Politécnica de Cartagena\\
 Spain\\ 
  \texttt{josel.sancho@upct.es} \\
\And
Aníbal R. Figueiras-Vidal\\
GAMMA-L+/DTSC\\ Universidad Carlos III de Madrid\\
 Spain\\
  \texttt{anibalrfv@tsc.uc3m.es} \\
\And
Francisco Herrera\\
 Andalusian Research Institute\\ in Data Science and\\ Computational Intelligence\\  University of Granada\\ 18071 
  Spain\\ 
  \texttt{herrera@ugr.es} \\
}

\begin{document}
\maketitle

\begin{abstract}
Ensemble methods have been widely used for improving the results of the best single classification model. A large body of works have achieved better performance mainly by applying one specific ensemble method. However, very few works have explored complex fusion schemes using heterogeneous ensembles with new aggregation strategies. This paper is three-fold: 1) It provides {\color{black} an overview} of the most popular ensemble methods, 2)  analyzes several fusion schemes using MNIST as guiding thread and 3) {\color{black} introduces MNIST-NET10, a complex heterogeneous fusion architecture based on a  degree of certainty aggregation approach; it combines two  heterogeneous schemes from the perspective of data, model and fusion strategy. MNIST-NET10 reaches a new record in 
 MNIST with only 10 misclassified images}. Our analysis shows  that such complex heterogeneous fusion architectures based on the degree of certainty can be considered as a way of taking benefit from diversity.

\end{abstract}

\section{Introduction}
In the last  decade, several types of Deep   Neural Networks (DNNs) have shown impressive results in extracting patterns from different data types. For instance, Convolutional Neural Networks (CNNs) constitute the state-of-the art in extracting patterns from images  \cite{krizhevsky2012imagenet}, while Recurrent Neural Networks with Long Short-Term Memory  units  constitute the state-of-the art in extracting patterns from text data   \cite{zaremba2014recurrent}. 

On the other hand, since the 70's, a large number of works in many fields has demonstrated that fusing several classifiers using a specific diversity strategy achieve a more stable and accurate global model with respect to the best individual one \cite{maimon2005data}. Given this high potential, there exist a large variety of ensemble methods, some train the learners in parallel on different partitions of the data, others train the learners sequentially on selected sets of samples and so on. In general, the more diverse is the ensemble, the better is the global model. Diversity can be introduced 1) through the involved base-learners in the ensemble; for example by using different architectures or training algorithms or 2) through the data by adopting a specific data-partition scheme or by using  different representations of the training samples or 3) by combining  both strategies. 

 The most adopted fusion schemes in the literature include one ensemble strategy. Very few works analyze complex fusion schemes that involve more than two ensemble methods. 
Since 1998, an important number of  these works have analyzed their approaches using the popular  handwritten digits classification problem with the  
 well known MNIST database \cite{lecun1998gradient}.

 MNIST was the first largest public dataset in  machine learning and since its creation it was utilized as a benchmark for evaluating different ensemble architectures. Several ensembles have been built to continuously reduce its test error rate. Currently, the most accurate fusion architecture misclassifies only 13 images over 10,000 test images \cite{R_4_3_7}.



In this paper, we do not intend to give an exhaustive overview of ensemble
methods as we are aware that there is a wide variety of these  fusion methods. Our aim is to provide a  {\color{black} brief overview}  of  fundamental ensemble methods in machine learning,  then focus our attention on the most used deep learning based fusion methods for MNIST digits classification as guiding thread. {\color{black}Following the fusion idea, we propose and analyze  MNIST-NET10, a  new complex fusion scheme  combined with a new aggregation technique that  reduces the  error rate in  MNIST  to 0.1$\%$.  We show that using such complex fusion schemes can  increase diversity and performance.}



The contributions of this paper can be summarized  as follows:
\begin{itemize}
    \item It presents  {\color{black}a brief overview} of  most popular  ensemble methods. 
    \item  It  provides the state-of-the art in MNIST with ensembles.
    \item {\color{black}It introduces MNIST-NET10, a  complex heterogeneous fusion architecture based on a new degree of certainty aggregation strategy, that reaches a new record in 
 MNIST with only 10 misclassified images. MNIST-NET10 combines two complex heterogeneous schemes from the perspective of data, learners architecture and fusion scheme. The  first scheme  is based on multiple well performing CNN architectures in combination with different data-augmentation techniques. The second  is a pipeline of two different ensemble methods, ECOC in the first stage and Bagging or Label Switching in the second stage in combination with CapsNet as data transforming model.
 
 Our experiments show that such  complex   heterogeneous fusion  schemes in combination with the degree of certainty aggregation technique can be used for getting benefit from diversity.}
\end{itemize}

This paper is organized as follows: An overview of fundamental ensemble methods is presented in  Section 2. Preliminaries and background to understand the contributions of this paper are given in Section 3.  The design and evaluation of the proposed complex fusion scheme and aggregation method are provided in Section 4, and finally, conclusions in Section 5. 

\section{Overview of most used ensemble  methods}
Ensemble learning consists of first creating  multiple predictive models then combing their predictions to obtain a better global model with a smaller
generalization error rate than the best single model. Ensembles  effectiveness increases with  learners  diversity~\cite{maimon2005data}, where diversity can be obtained by using different partitions or representations of the original training data and/or different designs of the learners, i.e.,  different  architectures, optimization algorithms, and hyper parameters.

When building an ensemble, several aspects should be taken into account: 
\begin{itemize}
    \item How diversity is introduced into the ensemble. Through data, through the model design or both. Ensembles can be homogeneous by including base learners  of the same type or  heterogeneous by including base learners  of different type. 
    \item The  number of learners.  In some cases, the number of learners  depends on the adopted ensemble strategy in others depends on the training-data partitions. For instance, One Versus One binarisation technique produces $n*(n-1)/2$ base learners, where $n$ is the number of object classes in the target problem. 
    \item  The order in which these learners are trained can be either sequential or  parallel. 
    \item The aggregation strategy to deduce the final prediction from the individual predictions can be 1)  a linear combination, e.g., sum, median, maximum, minimum or weighted sum functions, 2) a voting strategy, e.g., majority voting, plurality voting, weighted voting or soft voting, 3) {\color{black}  meta-combiner, which trains another learner, called meta-model, on  the predictions of the base learners, specific examples of this strategy include stacking and blending (see description below)}.
    
    
\end{itemize}

The most used ensemble strategies are bagging, boosting, label switching, mixture of experts, {\color{black} stacking, blending, } comibination of well performing models and binarization techniques.



\begin{itemize}
    \item[] \noindent\textbf{Bagging} (\underline{B}ootstrap \underline{agg}regat\underline{ing}) \cite{breiman1996bagging}  introduces diversity through data. The learners of the same design are trained in parallel  on  bootstrapped versions of the original training data, created using  extraction with replacement sampling. The final prediction of the ensemble is  calculated usually by averaging the values or using a majority vote.


    \item[] \textbf{Boosting}, also known as Adaptive Resampling \cite{freund1999short}, 
    in its basic form, each instance in the training dataset has a weight according to the previous results. The successive classifiers are generated by increasing the weight of the instances that are not predicted correctly and decreasing the weight of the instances that are correctly predicted. Each classifier specializes in the difficult instances for the previous classifier. This weight mechanism is also called pre-emphasis.
    
    Boosting was used for the first time with CNNs on MNIST in \cite{lecun1998gradient} and it reduced the test error of LeNet of MNIST from $0.8\%$ to $0.7\%$.

    \item[]\noindent\textbf{Label switching} (LS)
     is  a variant of the output flipping strategy proposed  in \cite{breiman2000randomizing}.  It introduces diversity by altering the labels of a proportion of the training samples using a stochastic mechanism. The simplest mechanism is based on a purely random selection. The selection of the switching rate depends on the data and the problem itself \cite{R_4_3_1}.

    \item[]\noindent\textbf{Mixture of experts}
     was first designed for problems whose data space can be divided into multiple homogeneous regions, because   the data was produced under different regimes \cite{jacobs1991adaptive}. Each learner becomes an expert on a sub-space by employing a special error function. In general, a supervised gating network is dedicated to combine the experts decisions  \cite{masoudnia2014mixture}.
    
    \item[]\noindent {\color{black} \textbf{Stacking and Blending}. In stacking method  the base learners are trained using a k-fold partition then the predictions are made on the left out fold. The predictions on this left out fold are then used to train another learner, called  meta model,  which will make the final prediction~\cite{Wol92}.  Blending  can be considered as a simple variation of stacking. Instead of creating out-of-fold predictions using k-fold, it creates a small hold-out data-set which will then be used to train the meta-model~\cite{sill2009feature}}

    \item[]\noindent\textbf{Combination of  well performing classifiers} consists of combining multiple well performing classification models that were trained on the entire sample space  considering that all the data points have equal weight \cite{tabik2017snapshot}. The final  decision of this ensemble is calculated either using a stacked model or using a simple voting approach.

    \item[]\noindent\textbf{Ensembles of multiple binary classifiers} converts  one multi-class classification problem into a number of binary classifiers and calculates the final prediction as a combination of  the predictions of the corresponding classifiers. The most known binarization approaches are  One-Versus-All (OVA), One-Versus-One (OVO) and Error Correcting Output Code (ECOC).
    
\begin{itemize}
    \item In OVA strategy \cite{clark1991rule, anand1995efficient}, each classifier learns how to distinguish each individual class versus all the rest of classes together. This approach produces as many classifiers as the number of classes in the original problem. The final prediction is obtained using an aggregation method called Maximum confidence strategy (MAX),  which considers the class with the largest output value as the predicted class. 




\item  OVO strategy \cite{knerr1990single, platt2000large, abe2003analysis} translates the original multi-class problem into as many binary problems as all the possible combinations between pairs of classes so that each classifier learns to discriminate between each pair. That is, a $m-$class problem will be converted into $m(m-1)/2$ classifiers. 
OVO can use diverse aggregation strategies. Namely, the Max-Wins rule (VOTE), Weighted Voting strategy (WV), Learning Valued Preference for Classification (LVPC), Preference relations solved by Non-Dominance criterion (ND), Classification by Pairwise Coupling (PC), Probability Estimates by pairwise coupling approach (PE) and Distance-based relative competence weighting combination for OVO (DRCW-OVO). 

\item 
ECOC binarization scheme assigns a unique binary string, also called codeword, to each class \cite{R_4_3_17}. These codewords are organized in a   table, in which each row represents one class by means of a binary codeword. The table has $n$ rows ($n$ is the number of classes) and $m$ columns that induce $m$ binary classification problems. Thus, $m$ binary classifiers are learned for each column. To classify a new data point, all $m$ binary classifiers are evaluated to obtain a $m$-bit string. Finally, the class (row in the table) whose codeword is closest to the $m$-length output string is chosen.
    According to this scheme, 15 binary classifiers are obtained for MNIST \cite{R_4_3_17}. To classify a new test image, all 15 binary classifiers are evaluated to obtain a 15-bit string. The predicted label is the class whose codeword is the closest to the output string. 
\end{itemize}
\end{itemize}

The  selection of the correct number of learners in ensembles has a significant impact on the performance of the global model. A low number of learners may cause  unstable classification accuracy, whereas a large number of learners  may increase the probability of redundant classifiers and hence result in less diversity.  Instead of combining all the learners,  pruning  approaches can be used to minimize the number of learners without losing generalization capacity \cite{galar2016ordering}. This approach is frequently used to achieve a  better trade-off between performance and computational cost.


\section{Background and Ensembles in MNIST digits  classification}

MNIST classification problem has been  addressed with many approaches related to data pre-processing, deep neural networks and ensemble methods. This section provides a brief description of all these aspects, data pre-processing (Section 2.1), deep learning (Section 2.2) and a summary of top-4 fusion schemes in MNIST digits classification (Section 2.3).

\subsection{Data pre-processing}

Data pre-processing is an essential element for the automatic learning process \cite{garcia2015data}. We can distinguish between two types of data pre-processing techniques. The first type is used to correct the data deficiencies that may affect the learning process, such as missing values, noise and outliers. The second type is used to simplify and optimize the training of the classification model by adapting the data,  modifying its dimensionality or increasing the number of training samples. The second type of pre-processing techniques is essential for supervised training of DNN models.

\begin{itemize}
    
\item[] {\bf Data augmentation:} As the process of manually constructing new labeled data sets is costly, in practice, data augmentation is used to increase the number of instances in the training dataset. It applies specific deformations, such as  translation and rotation, to the samples of the training dataset without altering the  existing spacial pattern in the original data \cite{tabik2017snapshot}. This type of pre-processing techniques makes the classification model more robust to these transformations in the images. 

In this work, for building and analyzing several deep learning ensembles in MNIST, we will consider the following data-augmentation techniques: rotation, translation, cropping, elastic deformation and Gaussian smoothing.

\item[] {\bf Transformation of data:}
Transforming models which produce cleaner representation of the input are also shown to be useful for improving robustness of DL models. Examples of these methods are Auto-encoders (AE) \cite{charte2018practical}, Denoising Auto-encoders (DAE) \cite{R_4_3_3,Brahma.2016} and Stacked Denoising Auto-encoders (SDAE) \cite{R_4_3_2}. CapsNet can be considered as a type of AE that can be used to create cleaner data with lower dimensionality \cite{R_4_3_6}. DAE  train a simple AE to reconstruct the input from a corrupted version of it,  obtained by  applying a stochastic corruption step  on the input.  Deeper representational models such as  SDAE are obtained by stacking multiple DAE layers. Each layer is trained as a  DAE by minimizing the error in reconstructing its input (which is the output code of the previous layer). 

In this work, to develop different ensemble methods for MNIST, we used  CapsNet   as data-transformer method as well as classification model.  We also used SDAE as a data transformer.

\end{itemize}

\subsection{Deep learning for handwritten digit classification}

In general, a CNN is built by staking three main building blocks, known as layers: 1) Convolutional layer, which is used to extract features at different levels of the
CNN hierarchy, 2) pooling layer, which is essentially a reduction operation used to increase the
 abstraction level of the extracted features, and 3) fully connected layer, which is used as a
 classifier at the end of the 
 pipeline. Under the supervised paradigm, CNN models are trained in an end-to-end fashion using a large labeled dataset.

Four of the most important architectures for  MNIST problem are  Network3~\cite{lecun1998gradient}, DropConnect~\cite{wan2013regularization}, MCDNN\cite{cirecsan2012multi}  and CapsNet \cite{R_4_3_6}.

\begin{itemize}
\item[] \noindent\textbf{Network3}  is based on the well known LeNet and it was presented in  \cite{Network3_2015}. Network3 consists of two convolutional layers (each one followed by max pooling)
and three fully connected layers with
Rectified Linear Unit (relu) as activation function. Cross entropy is used as loss function and
the output layer consists of 10 neurons with softmax
activation function. Both architectures have been
trained using the SGD algorithm.
\begin{table}[H]
 \centering
 \footnotesize{
 \begin{tabular}{|l|c|c|c|}
  \hline     Layer & Filter size & Stride & Activation\\ \hline
     \hline     conv1 & $5 \times 5 \times 20$ &  1 & -- \\
    maxpool1 & $2 \times 2$  & 2 & -- \\
    conv2 & $5 \times 5 \times 40$ &  1& relu \\
    maxpool2 & $2 \times 2$ &  2 & -- \\
    fc1 & $500$  & -- & relu \& Dropout rate =0.5 \\
    fc2 & $1000$ & -- & relu \& Dropout rate =0.5 \\
    fc3 & $10$ & -- & SoftMax \\\hline
 \end{tabular}}
 \caption{\label{tab:errors-dataaugmentation}Topology of Network3.}
\end{table}

\item[] \noindent\textbf{DropConnect} network has a similar architecture as Network3. The main difference is that it applies DropConnect instead of Dropout as regularization technique to the first fully connected layer~\cite{wan2013regularization}. {\color{black} DropConnect technique can be considered as a generalization
of Dropout  \cite{hinton2012improving}, for regularizing large fully connected layers. During the training process Dropout sets a randomly selected subset of neurons or nodes  to zero; while DropConnect  sets a randomly selected subset of weights within the network to zero.}

\begin{table}[H]
 \centering
 \footnotesize{
 \begin{tabular}{|l|c|c|c|}
  \hline     Layer & Filter size & Stride & Activation\\ \hline\hline
conv1 & $5 \times 5 \times 32$ & 1 & -- \\
maxpool1 & $2 \times 2$  & 2 & -- \\
conv2 & $5 \times 5 \times 64$ & 1& -- \\
maxpool2 & $2 \times 2$ &  2 & -- \\
fc1 & $150$ & -- & relu \& DropConnect rate:  0.5 \\
fc2 & $10$ & -- & softMax \\
\hline
 \end{tabular}}
 \caption{\label{tab:errors-dataaugmentation}Topology of DropConnect network.}
\end{table}

\item[] \noindent\textbf{MCDNN} combines several DNN columns to form a Multi-column DNN (MCDNN) \cite{cirecsan2012multi}. This architecture achieved a test error $0.23\%$ using diverse data pre-processing and a specific MCDNN architecture. In particular, they created six datasets from the original MNIST by normalizing the digits from $28\times28$ to  $10\times10$, $12\times12$, $14\times14$, $16\times16$, $18\times18$, and $20\times20$  pixels. Then they trained five DNN columns per normalization, resulting in a total of 35 columns for the entire MCDNN. Each DNN has a convolutional layer with 100 maps and 5x5 filters, a max-pooling layer over non overlapping regions of size 2x2, a convolutional layer with 40 maps and 5x5 filters, a max-pooling
layer over non overlapping regions of size 3x3, a fully connected layer with 150 hidden units and, a fully connected layer  with 10 neurons (one per class). The scaled hyperbolic tangent activation function is used for convolutional and fully connected layers, a linear activation function for max-pooling layers and a softmax activation function for the output layer. Each DNN column is trained using on-line gradient descent with an annealed learning rate. During training, images are continually translated, scaled and rotated (even elastically distorted in case of characters), whereas only the original images are used for validation. The learning  rate is initialized with 0.001 multiplied by a factor of 0.993/epoch until it reaches 0.00003.

\item[] \noindent\textbf{Capsule Networks (CapsNet)} 
was proposed to avoid the destruction of information produced by the max pooling operation \cite{R_4_3_6,R_4_3_7}.  This architecture  contains two capsule layers, which  are nonlinear computational elements whose inputs and outputs are vectors instead of scalar values. Using dynamic routing training algorithms, the probability and {\color{black} the  information included in an input image} (an object, for example) are coded by the length and the direction of the corresponding vector. This makes CapsNets equivariant, i.e., they are invariant to the point of view of the image. Such  property means that a CapsNet can identify new, unseen variations of the class images without ever being trained with them. Other forms of dynamic routing algorithms are presented in \cite{R_4_3_14, R_4_3_15}. \textcolor{black}{Figure \ref{CapsNet} shows a scheme of CapsNet network. }

\begin{figure}[h!]
         \centering
         \includegraphics[width=0.5\textwidth  ]{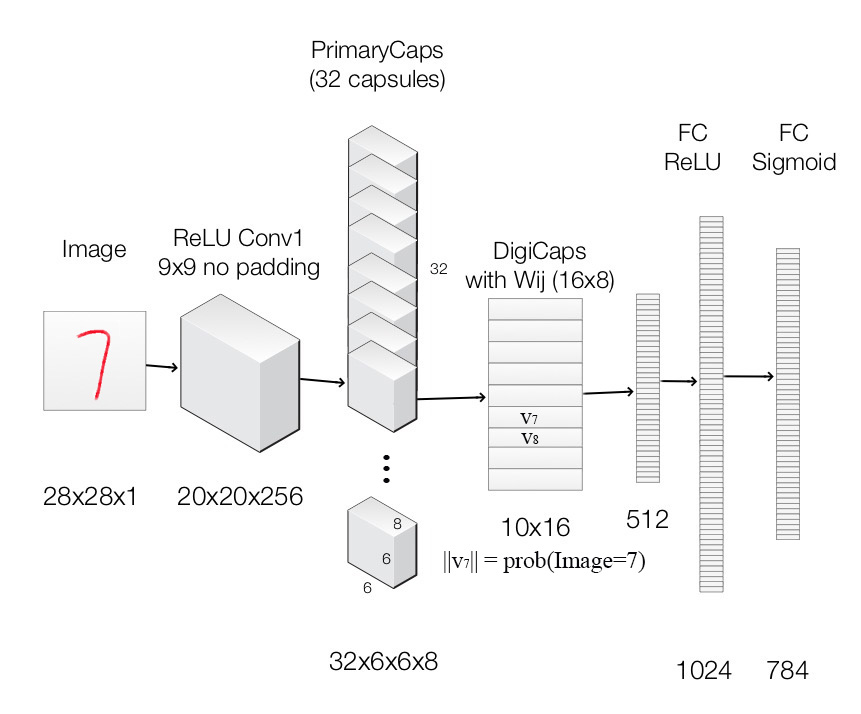}
         \caption{\color{black} An illustration of CapsNet architecture.}
\label{CapsNet}
\end{figure}

This network is actually an encoder-decoder network with the topology described in Table 3.

\begin{table}[H]
 \centering
 \footnotesize{
 \begin{tabular}{|l|c|c|c|c|c|c|}
 \hline     
        & Layer         & Input                     & Output                    & Capsules          & Kernels                   & Activation    \\
        &               &                           &                           & \#/size         & \#/size                 &               \\ \hline\hline
Encoder & Conv          & $28\times28$                & $20\times 20\times256$      & --                & $256/(9\times9)$          & ReLU          \\
        & PrimaryCaps   & $20\times20\times256$       & $6\times6\times8\times32$   & $1152/(8\times1)$ & $256/(9\times9\times256)$ & --            \\
        & DigitCaps     & $6\times6\times8\times32$   & $16\times10 $               & $10/(16\times1)$         & --                        & Squash        \\ \hline
Decoder & FC            & $16\times10$                & $512\times1$                & --                & --                        & ReLU          \\
        & FC            & $512\times1$                & $1024\times1$               & --                & --                        & ReLU          \\
        & FC            & $1024\times1$               & $784\times1$                & --                & --                        & Sigmoid       \\
\hline

 \end{tabular}}
 \caption{\label{tab:errors-dataaugmentation}Topology of CapsNet. {\color{black} Label "\#$/$size" refers to the number of capsules and their dimension.}}
\end{table}

\end{itemize}

\subsection{State-of-the art of fusion methods for MNIST digits classification}
 Since 1998, a large variety of fusion methods combining up to two ensemble strategies and different  diversity methods have been used to further reduce the error rate in MNIST. Currently, the  top four accurate ensembles for classifying handwritten digits are as follows. 

 \begin{itemize}
 \item The top-4 ensemble achieves a test error of $0.21\%$ {\color{black} (i.e., 21 misclassified images)}  by using five well performing CNN models \cite{wan2013regularization}. The CNN base learners are  based on the same architecture called  DropConnect network  and were trained on different sequences of random permutations of the training data. As pre-processing, the data was augmented by 1) randomly selecting cropped regions from the images, 2) flipping images horizontally, 3) introducing 15\% scaling and rotation variations. The final prediction is obtained using the most voted strategy.

\item  The top-3 model \cite{R_4_3_1} used a two-level ensemble which achieves an average test error of $0.19\%$ {\color{black} (i.e., 19 misclassified images)} . In the first level,   the Error Correcting Output Code (ECOC) binarization technique described in\cite{R_4_3_17} was used in combination with a pre-emphasis weighting strategy. SDAEs \cite{R_4_3_2} were used as base learners  trained  using  elastic deformation as data augmentation technique.  In a second level, a bagging ensemble is applied and the final decision is calculated according to a max vote.

\item The top-2 ensemble  achieved an error rate of $0.17\%$ ({\color{black} (i.e., 17 misclassified images)} ) \cite{loquercio2017computational}. The authors combined seven CNNs, from CNN0 to CNN6, and fifteen traditional classifiers, e.g., Random Forest and K-Nearest Neighbors.  CNN0, CNN2, CNN3, and CNN4 were used as feature extractors. The generated predictions are finally combined using a unsupervised meta-classifier called meta-Net based on the Einstein sum defined in \cite{buscema2013meta}. The used data-augmentation techniques were not specified in the paper.

\item The top-1 ensemble   achieved an average error rate of $0.13\%$ ({\color{black} (i.e., 13 misclassified images)} ) \cite{R_4_3_7} by first employing ECOC binarization technique described in \cite{R_4_3_17} combined with a pre-emphasis weighting strategy. The authors used the elastic deformation  as data augmentation technique to train the CNN learners. The output of the CNNs was processed using a SDAE \cite{R_4_3_2} to produce a new representation of the data. Then, finally,  a  Bagging ensemble was applied.

\end{itemize}

\section{Complex  fusion schemes for MNIST}

In this section, we introduce and evaluate MNIST-NET10, a  complex heterogeneous fusion architecture based on a new  aggregation strategy named degree of certainty aggregation method. In particular,  we first describe all the data pre-processing used in the experiments (Subsection 4.1).  We then evaluate two different fusion designs. The first design is based on multiple well performing CNN architectures in combination with different data-augmentation techniques (Subsection 4.2). The second design is a pipeline of two different ensemble methods, ECOC in the first stage and Bagging or Label Switching in the second stage in combination with CapsNet as data transforming model (Subsection 4.3). Finally, we combine the best fusion schemes based on the new aggregation strategy into a complex heterogeneous one called, MNIST-NET10 (Subsection 4.4). 


The experimental analysis uses the standard  division of MNIST database  $50,000 / 10,000 / 10,000$  for training, validation and test respectively. The final results are calculated by averaging 50 runs, {\color{black} to ensure a stable standard variation value}. 

In this section, we first provide a description of all the used pre-processing techniques,  then we present an analysis of the two analyzed fusion schemes.

\subsection{Dataset preprocessing}
To evaluate the considered fusion architectures, we built seven datasets, labeled as Dataset-1, -2, -3, -4   -5, -6 and -7  using different combinations of data pre-processing techniques as follows:

\begin{itemize}
    




\item  Dataset-1 was built by applying four random rotations followed by random elastic deformations then random rotations using the same parameters as in  the previous cases. The obtained dataset is nine times larger than the original one.

\item Dataset-2 was built by applying  random horizontal and vertical pixel translations plus a Gaussian smoothing. The selected parameters  for the translation and the Gaussian variance do not alter the digits visually. The obtained dataset is  $4\times$ larger than the original dataset.

\item {\color{black} Dataset-3 is  obtained from  Dataset-2 using   the pre-trained CapsNet available in this link: https://github.com/Sarasra/models/tree/master/research/capsules).  Each  $28\times 28$ input image of the original MNIST database is transformed into a $784\times 1$ array. It is worth noting that both encoder and decoder parts are used in this pre-processing.}

\end{itemize}

Table \ref{tab:errors-dataaugmentation} summaries the employed data pre-processing techniques  to create Dataset-1, 2 and 3.

\begin{table}[H]
 \centering
 \footnotesize{
 \begin{tabular}{|l|c|c|}
 \hline
 Dataset   & Pre-processing  & size w.r.t original \\\hline\hline
 Original  & -  & $1\times$  \\\hline
 Dataset-1 & rotation + elastic deformation & $9\times$ \\\hline
 Dataset-2 & translation + Gaussian smoothing & $4\times$ \\\hline
 Dataset-3 & {\color{black} Dataset-2 pre-processed using CapsNet} & $4\times$ \\
 
   &  {\color{black}Dataset-3 =  CapsNet(Dataset-2)}   &  \\\hline

 \end{tabular}}
 \caption{\label{tab:errors-dataaugmentation}The datasets created by applying different data preprocessing techniques to the original dataset.}
\end{table}

\subsection{A weighted fusion scheme based on  well performing classifiers and data-augmentation}
\label{Section_4_2}

We selected a set of well performing networks on MNIST: DropConnect\_1 and DropConnect\_2 \cite{cirecsan2012multi} (trained on dataset-1 and -2 respectively) and Network3 \cite{Network3_2015}, MCDNN \cite{wan2013regularization} and  CapsNet \cite{R_4_3_6} (trained on the original dataset). 
The performance of these models is shown in Table \ref{tab:errors-One}.

\begin{table}[H]
 \centering
\footnotesize{
\begin{tabular}{|l||c|}
     \hline
     Network & test error ($\%$)\\\hline
     \hline
     DropConnect\_2 & 0.18 \\
     DropConnect\_1 & 0.25 \\
     CapsNet & 0.24 \\
     MCDNN & 0.28 \\
     Network3 & 0.56 \\
\hline
\end{tabular}}
\caption{\label{tab:errors-One} Test error obtained by the five analyzed CNNs: Network3, CapsNet and MCDNN  (trained on the original dataset), and  DropConnect\_1 and DropConnect\_2   (trained on dataset-1 and -2 respectively).}
\end{table}

Afterwards, we  analyzed different combinations of these five well performing networks  and found that  fusion schemes  {\bf F1}, CapsNet$|$MCDNN$|$DropConnect\_2, and {\bf F2}, CapsNet$|$MCDNN$|$DropConnect\_1$|$DropConnect\_2, provide the best results. Symbol $|$ indicates that the involved models are run in parallel. We also analyzed the fusion scheme labeled as {\bf FS1} by aggregating {\bf F1},  {\bf F2} and  Dropconnect\_2 using the majority vote. Notice that in this new fusion Dropconnect\_2, MCDNN, CapsNet and Network3 has  a weight of 3, 2, 2 and 1 respectively. As shown in Table \ref{tab:errors}, {\bf FS1}
reduces the  test error to $0.12\%$. The twelve misclassified  images are shown in Figure \ref{Figure_5_1}(a). {\color{black} The codes of the five well performing base learners are provided through this link: https://github.com/SihamTabik/MNIST-NET10}.

       
\begin{table}[H]
 \centering 
 {%
 \begin{tabular}{|l|c|c|}
 \hline
Fusion   & Fusion structure     &  Test  \\
scheme    &   &     error in $\%$\\\hline\hline

{\bf F1}  &  CapsNet$|$MCDNN$|$DropConnect\_2   & 0.14 \\\hline

{\bf F2}   & CapsNet$|$MCDNN$|$ DropConnect\_1$|$DropConnect\_2$|$ Network3 &  0.14 \\\hline
{\bf FS1}  & F1$|$ F2$|$ Dropconnect\_2     & 0.12 \\\hline
\end{tabular}
}
\caption{\label{tab:errors}Test error of fusion schemes, {\bf F1}, {\bf F2}, and  {\bf FS1}. Symbol $|$  indicates  that the  involved models are run in parallel. }
\end{table}

\subsection{A multi-level fusion scheme with heterogeneous ensemble methods}
\label{Section_4_3}

In this section, we evaluate a more sophisticated fusion scheme that   combines multiple heterogeneous ensemble methods with data transformation as pre-processing technique. In particular, inspired from the top-1  fusion architecture \cite{R_4_3_1}, we evaluated two fusion designs, {\bf FS2} and {\bf FS3}, as described below. For simplicity, we express their architecture using  symbols $|$  and $\rightarrow$ to indicate that the involved strategies  in the same ensemble level are applied in parallel or in serial respectively. The base learners in these fusions are actually a MultiLayer Perceptron (MLP) with one hidden layer. As pre-processing technique we used CapsNet to transform Dataset-2 into Dataset-3. {\color{black} All the needed codes to build fusion schemes, {\bf FS2} and {\bf FS3}, are provided through this link https://github.com/SihamTabik/MNIST-NET10}.

\begin{figure}[h!]
     \centering
     \begin{subfigure}[b]{0.38\textwidth}
         \centering
         \includegraphics[width=\textwidth  ]{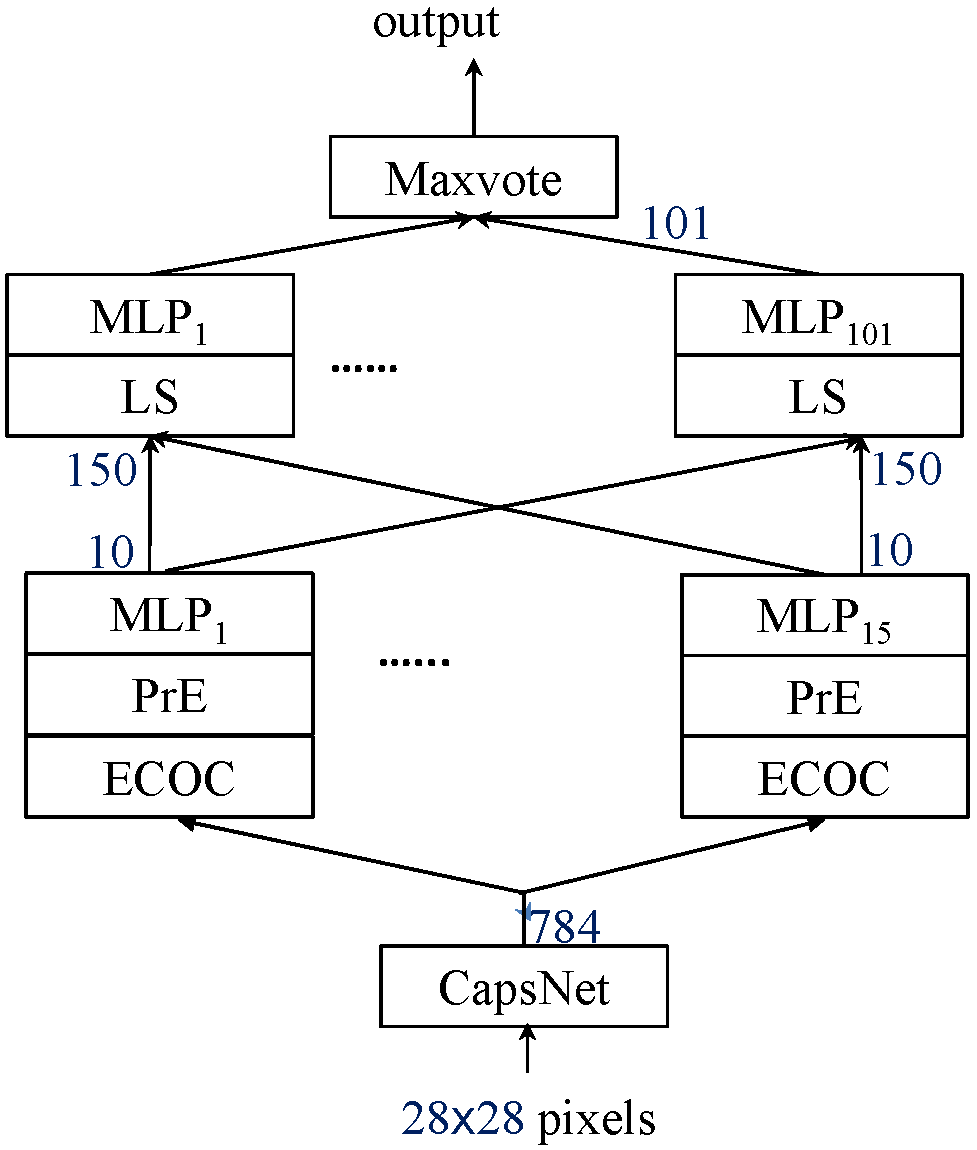}
         \caption{\bf FS2}
     \end{subfigure}
     \hspace{2em}
     \begin{subfigure}[b]{0.38\textwidth}
         \centering
         \includegraphics[width=\textwidth ]{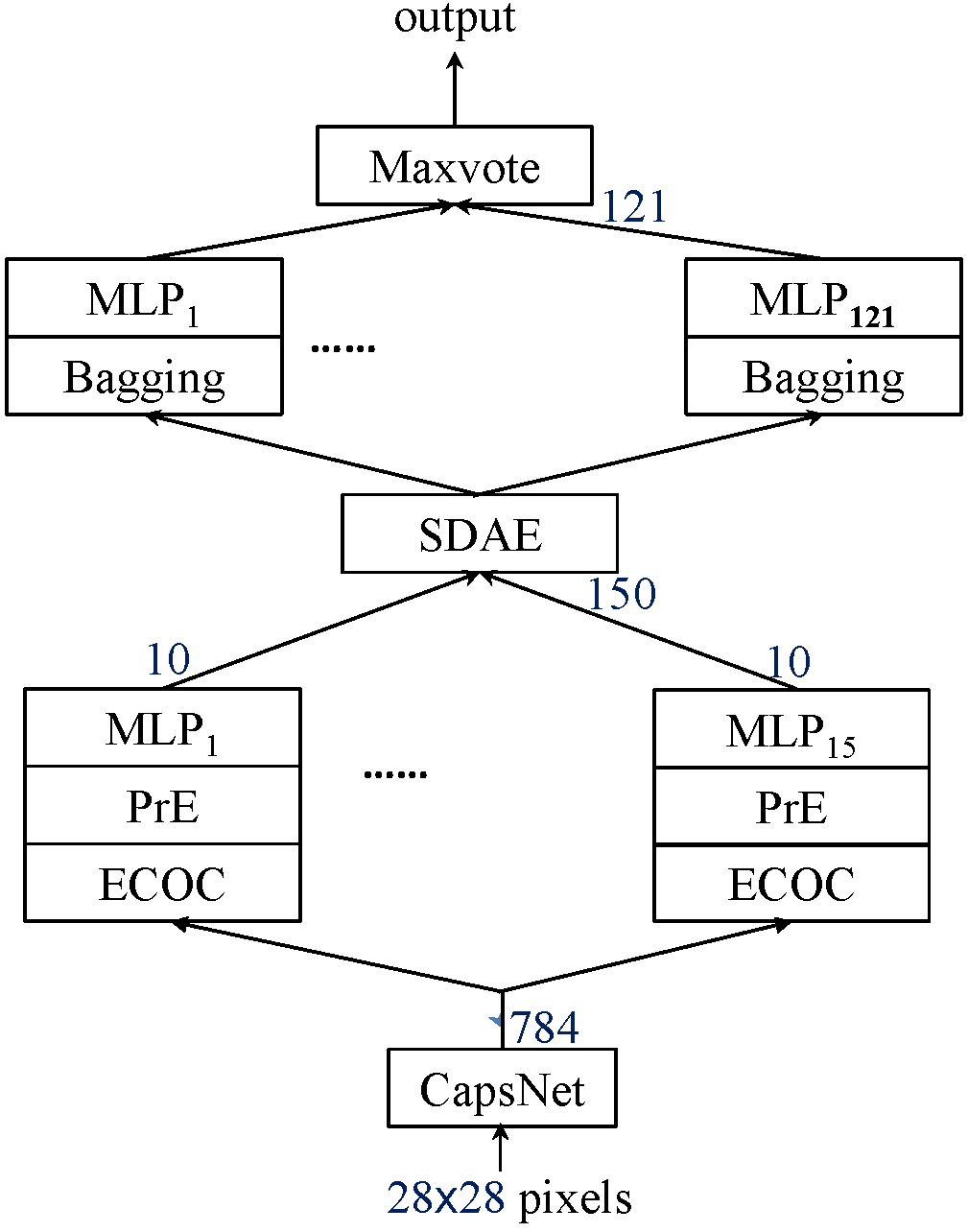}
         \caption{\bf FS3}
     \end{subfigure}
     \hfill
\caption{\color{black} An illustration of fusion schemes {\bf FS2} and  {\bf FS3} when trained/tested on the transformed digits with CapsNet. The dimension of the input and output is indicated at each level (in dark blue color).}
\label{Figure_FS2_FS3}
\end{figure}

\begin{itemize}

\item[] {\bf FS2}: ECOC$|$PrE$|$MLP$\rightarrow$LS$|$MLP, the  base-learners of the first level in this  scheme  are trained on Dataset-2. An adjustable weighting of the training samples, or Pre-Emphasis (PrE), is separately applied to each dichotomy as follows:
\begin{equation}
w(\mathbf{x^{(n)}})= \alpha+(1-\alpha)[\beta (t^{(n)}-o^{(n)})^2+(1-\beta)(1-o^{ (n)2})]
\end{equation}
where $\mathbf{x^{(n)}},t^{(n)} \in \{ -1,1 \}$, and $o^{(n)} \in [ -1, 1 ]$ are the training samples, its target and the corresponding {\color{black} model prediction}, respectively, and $\alpha, \beta \in [ 0,1]$ {\color{black} are convex combination parameters obtained by means of a grid 5-fold cross-validation process}.  Note that there is a  constant term, a term which is proportional to the square of the classification error, and a term which emphasizes the proximity to the classification border. 

This weighting form is a block version of the adjustable emphasis weights that {\color{black} were introduced in \cite{R_4_3_18} and successfully applied to construct boosting ensembles \cite{R_4_3_19, R_4_3_20}}.

Label Switching (LS) was used as diversity mechanism  in the second level, where the considered switching rate (S) and the number of classifiers (M) are selected among $\{ 10\%, 20\%, 30\%, 40\% \}$ and $\{ 11, 21, 51, 101\}$, respectively,  according to the classification performance with the validation set for each dichotomy.

\item[] {\bf FS3}: ECOC$|$PrE$|$MLP$\rightarrow$Bagging$|$SDAE$|$MLP, the base-learners of the first level in this  ensemble are trained using Dataset-3, which was produced by CapsNet previously trained on MNIST\footnote{pre-trained CapsNet weights were obtained from: https://github.com/Sarasra/models/tree/master/research/capsules}.

 The architecture of  SDAE includes   three layers of $1000$ units each after a compression layer of $960$ units, the final classification is obtained by a bagging  ensemble of one hidden ($1000$ units) layer, selecting the bootstrap population sizes (B) and number of learners (M) among $\{ 60\%, 80\%, 100\%, 120\% \}$ and $\{ 21, 51, 101, 121\}$, respectively, according to the classification results in the validation set. {\color{black} As expected, there is a saturation effect since after a certain number of Switching (in FS2) or bagging (in FS3) ensembles there is no variation in the results and hence it is not necessary to explore a larger number.}

\end{itemize}

The final decision in both  fusion designs is taken by the majority
of the highest outputs of all these networks. The intrinsic redundancy of the resulting ensemble of binary problems produces improvements when applying minimum Hamming distance classification. \textcolor{black}{ Fusion schemes FS2 and FS3 are depicted in Figure \ref{Figure_FS2_FS3}.}

\begin{table*}[h]
\begin{center}
\renewcommand{\arraystretch}{1.25}
\begin{tabular}{|c|c|c|}
   
\hline
 Fusion      & \multicolumn{2}{c|}{\footnotesize Training Dataset}  \\  \cline{2-3}
  Scheme& \footnotesize{ Dataset-2} & \footnotesize{Dataset-3} \\
  
  \hline\hline
   {\footnotesize  {\bf FS2:}ECOC$|$PrE$|$MLP$\rightarrow$LS$|$MLP} & \footnotesize{ $0.14 \pm 0.01$}  &  \footnotesize{$0.12 \pm 0.00$}    \\  \hline
   {\footnotesize {{\bf FS3:}}ECOC$|$PrE$|$MLP$\rightarrow$Bagging$|$SDAE$|$MLP  } & \footnotesize{$0.13 \pm 0.00$} & \footnotesize{$0.12 \pm 0.02$}  \\
 \hline

\end{tabular}
\end{center}

\caption{Test error rates, expressed in $\%$ average $\pm$ standard deviation, for fusions schemes {\bf FS2}  and {\bf FS3}. Note that Dataset-3 is obtained by applying CapsNet as transforming model to Dataset-2}
 \label{Table_4_3_1}
\end{table*}

Table \ref{Table_4_3_1} presents the experimental results, $\%$ test average error rate $\pm$ standard deviation, for the two considered ensemble designs, {\bf FS2}  and {\bf FS3}. 

It has been checked that validation performances saturate for the extreme values of the switching and bagging ensemble sizes (M=$101$ and M=$121$, respectively). 

As it can be observed, using the  representations produced by CapsNets, i.e., Dataset-3, provides slight but clear and consistent improvements with respect to the MNIST input counterpart, i.e., Dataset-2, in both fusion schemes {\bf FS2} and {\bf FS3}. It must be highlighted that replacing the LS layer by  Bagging$|$SDAE  increases the  effectiveness of {\bf FS3} with respect to {\bf FS2} when using Dataset-2 but does not affect the performance of {\bf FS3}  on Dataset-3. This seems to be a consequence of the highest capacity of  CapsNet in extracting information from the original data, which makes useless the attempt of extracting more information at the second level of the ensemble by stacking  SDAE and Bagging.

\subsection{MNIST-NET10: A heterogeneous deep learning networks fusion based on the degree of certainty}

In this section, we introduce and evaluate MNIST-10, a heterogeneous  fusion scheme in combination with a  aggregation method, called  the degree of certainty aggregation approach.

\begin{figure}[h!]
     \centering
     \begin{subfigure}[b]{0.46\textwidth}
         \centering
         \includegraphics[width=\textwidth  ]{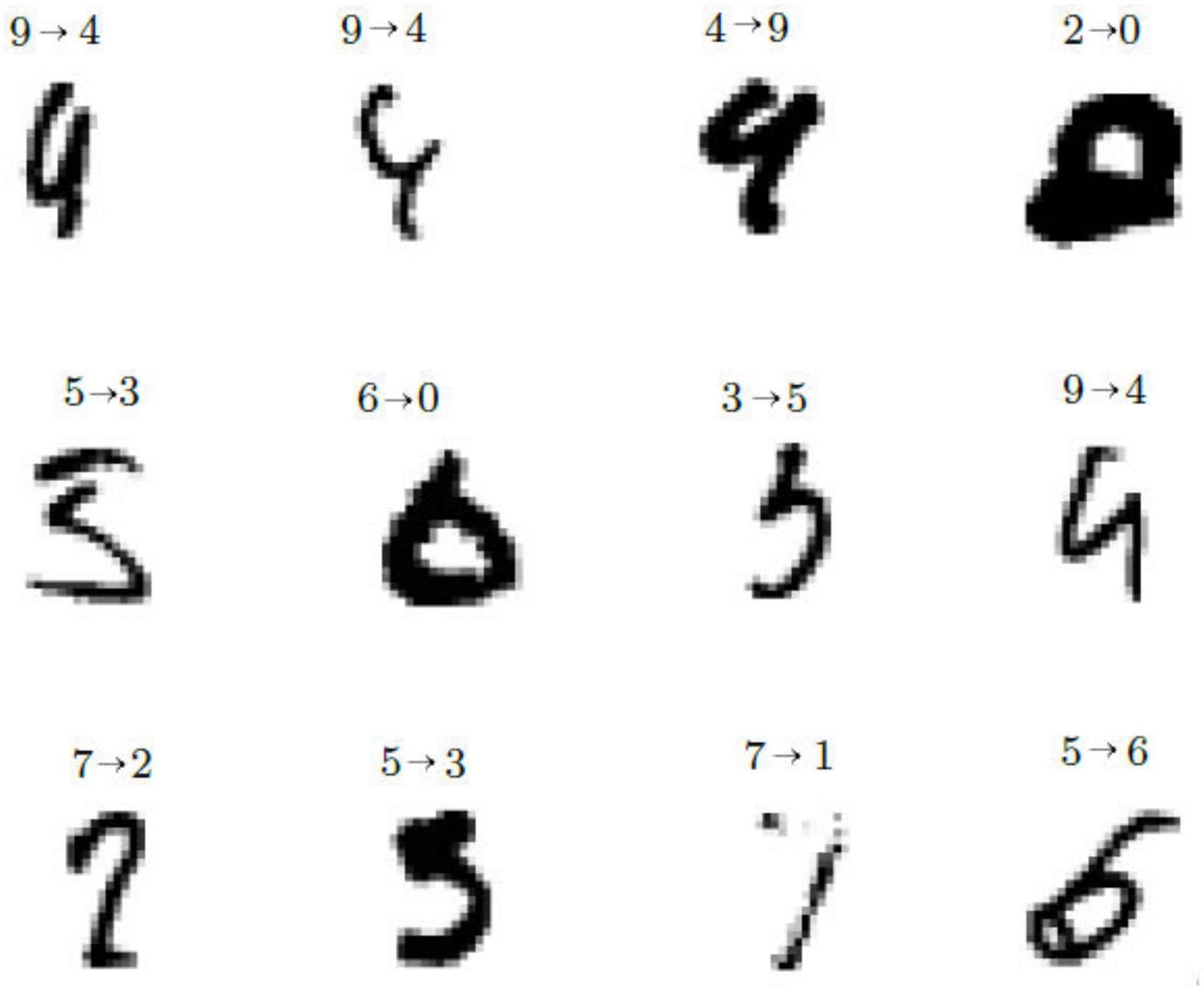}
         \caption{ }
     \end{subfigure}
     \hfill
     \begin{subfigure}[b]{0.46\textwidth}
         \centering
         \includegraphics[width=\textwidth ]{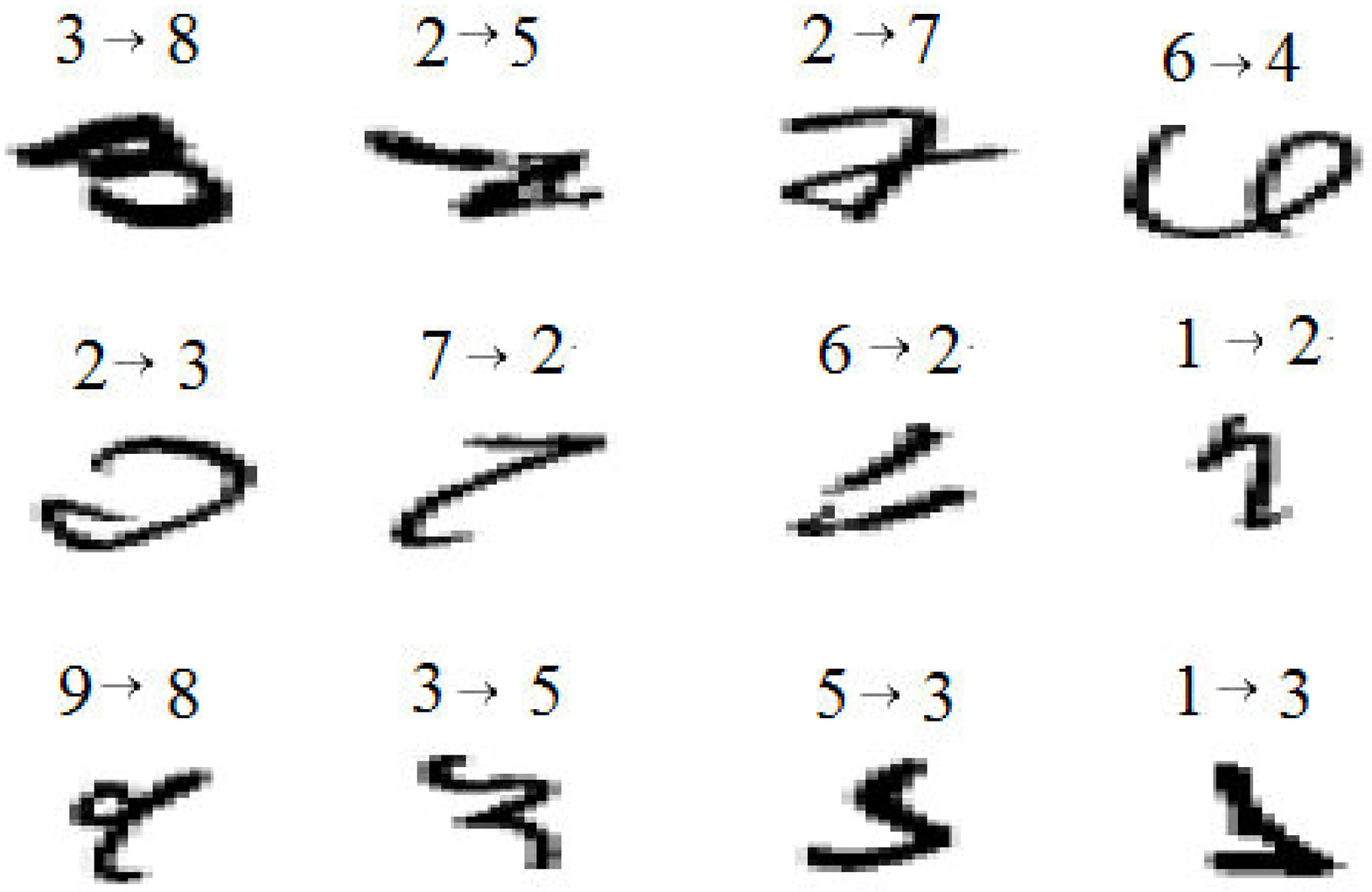}
         \caption{ }
     \end{subfigure}
     \hfill
\caption{Misclassified test digits (label $\rightarrow$ results) by (a) a run of ensemble {\bf FS1} (Subsection \ref{Section_4_2}) and (b) a run of ensemble {\bf FS2} with CapsNet input (Subsection \ref{Section_4_3}).  }
\label{Figure_5_1}
\end{figure}

The fusion schemes  presented in Subsections \ref {Section_4_2} and \ref {Section_4_3}  are heterogeneous from the perspectives of data, learners architecture and also fusion scheme. Indeed, as it can be observed from Figure \ref{Figure_5_1},  the misclassified test digits by each fusion design are completely different. This finding encouraged us to explore whether aggregating both ensemble outputs would produce improvements. 

We propose MNIST-NET10 by aggregating the best ensembles, {\bf FS1} from section \ref{Section_4_2} and {\bf FS2}  from section \ref{Section_4_3}. A reasonable and easy to implement aggregation strategy is as follows. If digits $d_1$ and $d_2$ are the winners for {\bf FS1} and {\bf FS2} respectively and 
$d_2 \neq d_1$, we compute the predicted digit of the combination of both ensembles, $d$, such that
$$d = \arg\max_{d_i} [ {v_1(d_i)/9+v_2(d_i)/101} ]$$

where  $v_j(d_i)$, with $i,j=1,2$, is the number of votes that $d_i$ receives at the output of the $j$th ensemble. This is an aggregation based on the degree of certainty of each fusion architecture for the candidate digits.  {\color{black} Values 9 and 101 corresponds respectively to the number of classifiers in   {\bf FS1}  and the number of classifiers in the last level of {\bf FS2}.}

\begin{figure}
\centering
\includegraphics[ width=0.6\textwidth]{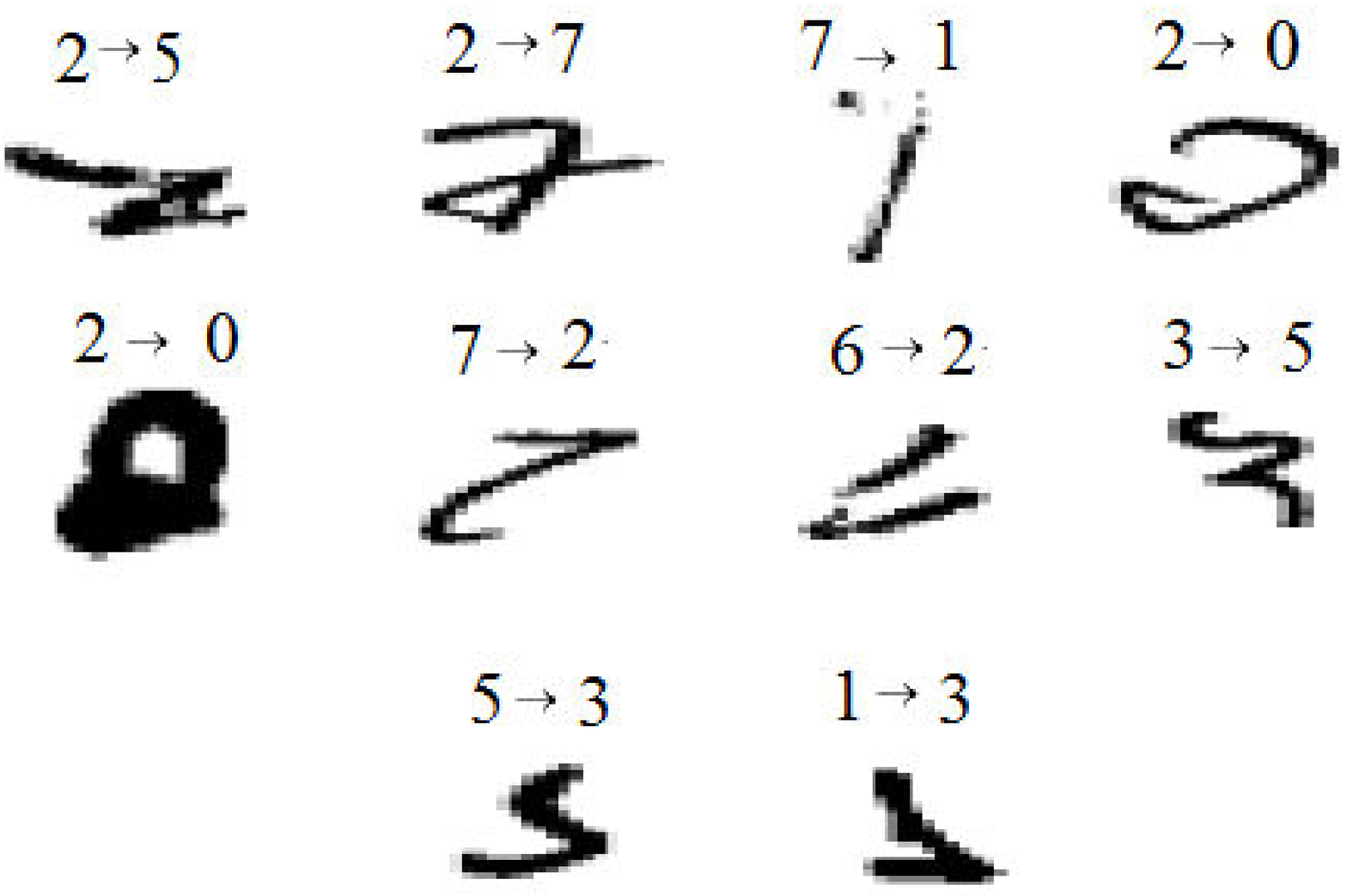}
\caption{Misclassified test digits  (label $\rightarrow$ result) obtained from  fusing {\bf FS1} and {\bf FS2}.}
\label{Figure_5_2}
\end{figure}

The performance result for 10 runs of this fusion scheme is $0.10\pm 0.02\%$, which is  better than the test error of the independent fusion schemes. The  misclassified digits for a run of the aggregated ensembles are shown in Figure \ref{Figure_5_2}. As we can observe from this figure, ten and four  misclassified digits respectively by  {\bf FS1} and {\bf FS2} from Figure \ref{Figure_5_1}(a) and (b) are now correctly classified by the fusion of both ensembles. However,  two and eight misclassified digits by  {\bf FS1} and  {\bf FS2} respectively are still misclassified by the fusion of both ensembles.

 It is interesting to note  that the remaining misclassified digits have strange shapes and are impossible to be correctly classified by a human eye. However, further exploiting the idea of combining heterogeneous ensembles could provide even (statistically) better results, although practical differences would be minor.

\section{Conclusions}

This paper introduces MNIST-NET10, a  complex heterogeneous fusion architecture based on a novel degree of certainty aggregation strategy, that achieves the state-of-the art accuracy in the problem of MNIST {\color{black}with only 10 misclassified images. Moreover, we present an overview of the most popular  ensemble methods and pre-processing techniques that when combined together can improve robustness and accuracy with respect to the best individual model. The state-of-that art  in MNIST with ensembles is also presented. }

As learnt lessons  from our study, we can state that building complex deep learning fusions,  by combining different heterogeneous ensemble methods considering deep learning neural networks, data augmentation and data transformation increases diversity and consequently increase robustness and efficiency of the global model. 

{\color{black}
In general, current research in deep neural networks can be split into two groups, the first group of studies focuses on further increasing performance of models while the second  focuses on improving explainability and accountability  of single classification  models \cite{arrieta2020explainable}. Improving performance by combining different forms of diversity with deep neural network designs is an important and still open area of research. Hence, exploring this large space of combinations is essential as it is plausible to expect excellent results in  many fields of application. In the mid and long term, explainability methods must be extended and adapted to the context of ensembles. 
}

\bibliographystyle{unsrt}  


\end{document}